\documentclass[10pt,twocolumn,letterpaper]{article}

\usepackage{cvpr}              

\usepackage{graphicx}
\usepackage{amsmath}
\usepackage{amssymb}
\usepackage{booktabs}
\usepackage{bm}
\usepackage{multirow}
\usepackage[accsupp]{axessibility}
\usepackage[pagebackref,breaklinks,colorlinks]{hyperref}

\usepackage[capitalize]{cleveref}
\crefname{section}{Sec.}{Secs.}
\Crefname{section}{Section}{Sections}
\Crefname{table}{Table}{Tables}
\crefname{table}{Tab.}{Tabs.}

\def\cvprPaperID{8250} 
\def\confName{CVPR}
\def\confYear{2023}

\def\papername{OmniAvatar}

\begin{document}

\title{OmniAvatar: Geometry-Guided Controllable 3D Head Synthesis}

\author{Hongyi Xu$^1$ \quad Guoxian Song$^1$ \quad Zihang Jiang$^{1,2}$ \quad Jianfeng Zhang$^{1,2}$ \quad Yichun Shi$^1$ \\  Jing Liu$^1$ \quad Wanchun Ma$^1$ \quad Jiashi Feng$^1$ \quad Linjie Luo$^1$\\
$^1$ByteDance Inc \quad\quad $^2$National University of Singapore\\
\tt\small \{{hongyixu, guoxian.song, zihang.jiang, jianfeng.zhang, yichun.shi,} \\
\tt\small{jing.liu, wanchun.ma, jshfeng, linjie.luo\}@bytedance.com}
}

\newcommand{\bb}[1]{\boldsymbol{\mathbf{#1}}}

\twocolumn[{
\renewcommand\twocolumn[1][]{#1}
\maketitle
\begin{center}
    \captionsetup{type=figure}
    \includegraphics[width=1.0\textwidth]{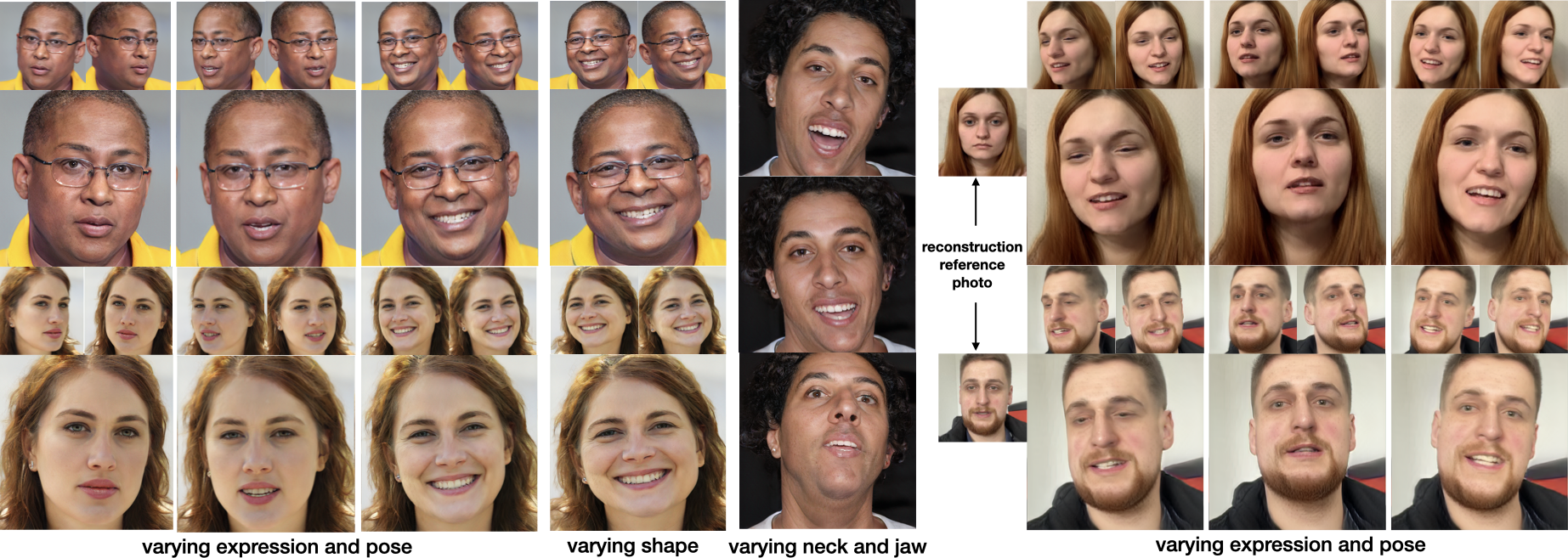}
    \captionof{figure}{Our model can synthesize diverse identity-preserved 3D heads with compelling dynamic details under full disentangled control over camera poses, facial expressions, head shapes, articulated neck and jaw poses (left). Our model can also reconstruct 3D heads from a single photo reference and enable multi-view-consistent head reenactment (right).}
    \label{fig:teaser}
\end{center}
}]


\begin{abstract}
We present \papername, a novel geometry-guided 3D head synthesis model trained from in-the-wild unstructured images that is capable of synthesizing diverse identity-preserved 3D heads with compelling dynamic details under full disentangled control over camera poses, facial expressions, head shapes, articulated neck and jaw poses. To achieve such high level of disentangled control, we first explicitly define a novel semantic signed distance function (SDF) around a head geometry (FLAME) conditioned on the control parameters. This semantic SDF allows us to build a differentiable volumetric correspondence map from the observation space to a disentangled canonical space from all the control parameters. We then leverage the 3D-aware GAN framework (EG3D) to synthesize detailed shape and appearance of 3D full heads in the canonical space, followed by a volume rendering step guided by the volumetric correspondence map to output into the observation space. To ensure the control accuracy on the synthesized head shapes and expressions, we introduce a geometry prior loss to conform to head SDF and a control loss to conform to the expression code. Further, we enhance the temporal realism with dynamic details conditioned upon varying expressions and joint poses. Our model can synthesize more preferable identity-preserved 3D heads with compelling dynamic details compared to the state-of-the-art methods both qualitatively and quantitatively. We also provide an ablation study to justify many of our system design choices.
\end{abstract}


\section{Introduction}
\label{sec:intro}

Photo-realistic face image synthesis, editing and animation attract significant interests in computer vision and graphics, with a wide range of important downstream applications in visual effects, digital avatars, telepresence and many others. With the advent of Generative Adversarial Networks (GANs)~\cite{goodfellow2014generative}, remarkable progress has been achieved in face image synthesis by StyleGAN~\cite{karras2019style,Karras2020stylegan2,karras2021alias} as well as in semantic and style editing for face images~\cite{shi2021SemanticStyleGAN,tov2021designing}. To manipulate and animate the expressions and poses in face images, many methods attempted to leverage 3D parametric face models, such as 3D Morphable Models (3DMMs)~\cite{blanz1999morphable,paysan20093d}, with StyleGAN-based synthesis models~\cite{deng2020disentangled,tewari2020stylerig,piao2021inverting}. However, all these methods operate on 2D convolutional networks (CNNs) without explicitly enforcing the underlying 3D face structure. Therefore they cannot strictly maintain the 3D consistency when synthesizing faces under different poses and expressions.

Recently, a line of work has explored neural 3D representations by unsupervised  training of 3D-aware GANs from in-the-wild unstructured images~\cite{chan2021pi,gu2021stylenerf,or2021stylesdf,xue2022giraffe,chan2021efficient,epigraf,xu2021generative,shi2021lifting,deng2021gram,schwarz2020graf,zhou2021CIPS3D}. Among them, methods with generative Neural Radiance Fields  (NeRFs)~\cite{mildenhall2020nerf} have demonstrated striking quality and multi-view-consistent image synthesis
~\cite{epigraf,chan2021efficient,or2021stylesdf,gu2021stylenerf,deng2021gram}. The progress is largely due to the integration of the power of StyleGAN in photo-realistic image synthesis and NeRF representation in 3D scene modeling with view-consistent volumetric rendering. Nevertheless, these methods lack precise 3D control over the generated faces beyond camera pose, as well as the quality and consistency in control over other attributes, such as shape, expression, neck and jaw pose, leave much to be desired.

In this work, we present \emph{\papername}, a novel geometry-guided 3D head synthesis model trained from in-the-wild unstructured images. Our model can synthesize a wide range of 3D human heads with full control over camera poses, facial expressions, head shapes, articulated neck and jaw poses. To achieve such high level of disentangled control for 3D human head synthesis, we devise our model learning \emph {in two stages}. We first define a novel \emph{semantic signed distance function} (SDF) around a head geometry (i.e. FLAME~\cite{FLAME:SiggraphAsia2017}) conditioned on its control parameters. This semantic SDF fully distills rich 3D geometric prior knowledge from the statistical FLAME model and allows us to build a differentiable \emph{volumetric correspondence map} from the \emph{observation space} to a disentangled \emph{canonical space} from all the control parameters. In the second training stage, we then leverage the state-of-the-art 3D GAN framework (EG3D~\cite{chan2021efficient}) to synthesize realistic shape and appearance of 3D human heads in the canonical space, including the modeling of hair and apparels. Following that, a volume rendering step is guided by the volumetric correspondence map to output the geometry and image in the observation space.



\begin{figure*}[t]
    \centering
    \includegraphics[width=1.\textwidth]{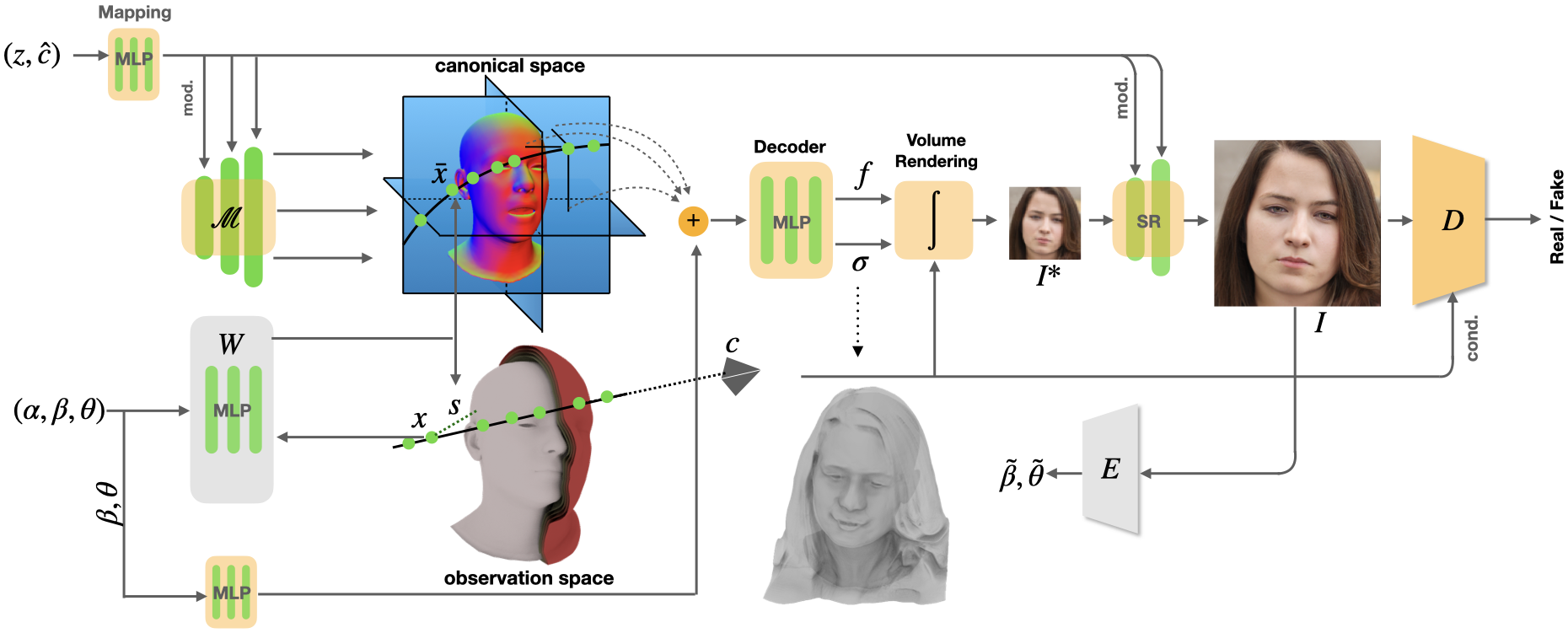}
    \caption{\textbf{Overview of our training framework.} \textbf{Stage 1}: Trained from parameterized FLAME~\cite{FLAME:SiggraphAsia2017} mesh collections, a MLP-network $W$ maps a shape  $\bb{\alpha}$, expression $\bb{\theta}$ and articulated jaw and neck pose $\bb{\theta}$ into 3D point-to-point volumetric correspondences from observation to canonical space, together with a signed distance function of the corresponding FLAME head. \textbf{Stage 2}: Given a Gaussian latent code $\bb{z},$ our model generates a tri-plane represented 3D feature space of a canonical head, disentangled with shape and expression controls. The volume rendering is then guided by the volumetric correspondence field to map the decoded neural radiance field from the canonical to observation space. We condition the NeRF decoding with expression and joint pose for modeling dynamic details. A super-resolution module synthesizes the final high-resolution RGB image from the volume-rendered feature map. For fine-grained shape and expression control, we apply the FLAME SDF as geometric prior to the synthesized NeRF density, and self-supervise the image synthesis to commit to the target expression $\bb{\beta}$ and joint pose $\bb{\theta}$ by comparing the input code against the re-estimated values $\hat{\bb{\beta}},\hat{\bb{\theta}} $ from synthesized images. }
    \label{fig:overview}
    \vspace{-0.1in}
\end{figure*}

To ensure the consistency of synthesized 3D head shape with controlling head geometry, we introduce a \emph{geometry prior loss} to minimize the difference between the synthesized neural density field and the FLAME head SDF in observation space. Furthermore, to improve the control accuracy, we pre-train an image encoder of the control parameters and formulate a \emph{control loss} to ensure synthesized images matching the input control code upon encoding. Another key aspect of synthesis realism is dynamic details such as wrinkles and varying shading as subjects change expressions and poses. To synthesize dynamic details, we propose to condition EG3D's triplane feature decoding with noised controlling expression. 

Compare to state-of-the-art methods, our method achieves superior synthesized image quality in terms of Frechet Inception Distance (FID) and Kernel Inception Distance (KID). Our method can consistently preserve the identity of synthesized subjects with compelling dynamic details while changing expressions and poses, outperforming prior methods both quantitatively and qualitatively. 


The contributions of our work can be summarized as:
\begin{itemize}
    \vspace{-0.08in}
    \item A novel geometry-guided 3D GAN framework for high-quality 3D head synthesis with full control on camera poses, facial expressions, head shapes, articulated neck and jaw poses.
    \vspace{-0.08in}
    \item A novel semantic SDF formulation that defines the volumetric correspondence map from observation space to canonical space and allows full disentanglement of control parameters in 3D GAN training.
    \vspace{-0.08in}
    \item A geometric prior loss and a control loss to ensure the head shape and expression synthesis accuracy.
    \vspace{-0.08in}
    \item A robust noised expression conditioning scheme to enable dynamic detail synthesis.
\end{itemize}

\section{Related Work}
\label{sec:relatedwork}

\paragraph{3D-Aware Generative Image Synthesis.}
Generative adversarial networks~\cite{goodfellow2014generative,karras2019style,Karras2020stylegan2} gained popularity over the last decade due to their remarkable ability in photo-realistic image synthesis. Building on the success of 2D image-based GANs, recent works have extended the capabilities to view-consistent image synthesis with unsupervised learning from 2D single-view images. The key idea is to combine differential rendering with 3D scene representations, such as meshes~\cite{Szabo:2019,Liao2020CVPR}, point clouds~\cite{achlioptas2018learning,li2019pu}, voxels~\cite{wu2016learning,hologan,nguyen2020blockgan}, and recently implicit neural representation~\cite{schwarz2020graf,chan2021pi,niemeyer2021giraffe,chan2021efficient,or2021stylesdf,gu2021stylenerf,deng2021gram,epigraf,zhou2021CIPS3D}. We build our work on recent 3D GAN model by Chan et al~\cite{chan2021efficient} that uses an efficient triplane-based NeRF generation, combined with 2D CNN-based super-resolution. Even though 3D-aware GANs are able to control camera viewpoints, they lack precise 3D control over the other attributes such as shapes and expressions. 
 In this work, we empower 3D-aware GANs with disentangled precise control over shapes and expressions. 
\vspace{-0.12in} 
\paragraph{Controllable Face Image Synthesis.}
Considerable work~\cite{deng2020disentangled,tewari2020stylerig,chen2022sofgan,piao2021inverting,tewari2020pie,kowalski2020config,ghosh2020gif} has been devoted to incorporate 3D priors of statistical face models, such as 3D Morphable Models (3DMMs)~\cite{blanz1999morphable,paysan20093d}, in controllable face synthesis and animation. Among them, DiscoFaceGAN~\cite{deng2020disentangled} proposed imitative-contrastive learning to mimic the 3DMM rendering process by the generative model. A similar strategy has also been adopted with concurrent and follow-up works~\cite{tewari2020stylerig,kowalski2020config,ghosh2020gif,piao2021inverting}. However all of these approaches suffer from 3D inconsistency due to the use of 2D CNNs as image renderer. 
HeadNeRF~\cite{hong2022headnerf} combines 3DMM with 3D NeRF representation, and is able to synthesize 3D heads conditioned on 3DMM attributes. However, the training relies on annotated multiview datasets whereas our approach learns the disentangled 3D head synthesis with only single-view images. There has been concurrent work~\cite{wu2022anifacegan,sun2022controllable,bergman2022generative,zhang2022avatargen,tang2022explicitly} to ours in 3D-aware controllable face or full-body GANs. Differently from these approaches, we use a full-head parametric model FLAME~\cite{FLAME:SiggraphAsia2017}, and fully exploit the spatial geometric prior knowledge beyond the surface deformation and skinning. We have also achieved fine-grained control with our novel losses and enhanced face animation with rich dynamic details. 
\vspace{-0.12in} 
\paragraph{Controllable Neural Implicit Field of Face.} 
Neural implicit functions~\cite{xie2022neural}, have emerged as a powerful continuous and differential representations of 3D scenes. Among them, Neural Radiance Field~\cite{mildenhall2020nerf,barron2021mip} has been widely adopted due to its superiority in modeling complex scene details and synthesizing multiview images with inherited 3D consistency. While initial proposals have focused on static scene modeling, recent work have successfully demonstrated application of NeRF in modeling dynamic scenes~\cite{pumarola2021d,park2021nerfies,xu2021h,liu2021neural,tretschk2021non}. In particular, dynamic neural radiance fields of human heads~\cite{park2021hypernerf,park2021nerfies,gafni2021dynamic,guo2021ad,wang2021learning,zheng2022avatar,kania2022conerf,zhuang2021mofanerf} have enabled photo-realistic head animation, often by conditioning with pose parameters or deforming the radiance field with 3D morphable models. However, they do not leverage a generative training paradigm and focus on scene-specific learning from video sequences or multiview images. In contrast, our model learns generative and controllable neural radiance fields from widely accessible single-view images.

\section{Method}

\label{sec:method}
Our goal is to build a geometry-guided 3D head synthesis model with full control of camera poses, face shapes and expressions, trained from in-the-wild unstructured image collections.

\subsection{Overview}
\paragraph{Problem.}
To achieve our goal, we leverage a 3D-aware generator (EG3D~\cite{chan2021efficient}) for photo-realistic, multiview consistent image synthesis, while disentangle control of head geometric attributes from image generation with a 3D statistical head model (FLAME~\cite{FLAME:SiggraphAsia2017}). Specifically, given a random Gaussian-sampled latent code $\bb{z}$, a camera pose $\bb{c}$ and a FLAME parameter $\bb{p}=(\bb{\alpha}, \bb{\beta},\bb{\theta})$ consisting of shape $\bb{\alpha},$ expression $\bb{\beta},$ jaw and neck pose $\bb{\theta}$, the generator $G$ synthesizes a photo-realistic human head image $I_{RGB}(\bb{z} | \bb{c}, \bb{p})$ with corresponding attributes as defined in $\bb{p}.$    
\vspace{-0.05in}
\paragraph{Framework.}
As illustrated in our pipeline~\ref{fig:overview}, 
our controllable 3D-aware GAN is trained in two stages. From a large collection of 3D deformed FLAME meshes, we first pre-train a deformable semantic SDF around the FLAME geometry that builds a differential volumetric correspondence map from the observation to a predefined canonical space (Section~\ref{sec:imflame}). In the second stage, guided by the pre-trained volumetric mapping, we then deform the detailed 3D full heads synthesized in the disentangled canonical space to the desired shapes and expressions (Section~\ref{sec:canonical}). Fine-grained expression control is achieved by supervising image synthesis such that expressions estimated from the generated images is consistent with the input control(Section~\ref{sec:supervision}). Our approach further enhances temporal realism with dynamic details, such as dimples and wrinkles, synthesizing realistic shading and geometric variations as expression changes (Section~\ref{sec:dynamic}).  

 
\vspace{-0.1in}
\paragraph{3D-Aware GAN Background.}
To ensure appearance consistency from different views, we choose EG3D~\cite{chan2021efficient} as our backbone for 3D-aware image synthesis. The generator $G$ takes a random latent code $\bb{z}$ and conditioning camera label $\hat{\bb{c}}$, and maps to a manifold of triplane features $\cal{M}(\bb{z}, \hat{\bb{c}})$. For presentation clarity, we absorb $\hat{\bb{c}}$ to $\bb{z}$ and simply denote triplane as $\cal{M}(\bb{z}).$
A low-resolution feature map $I^{*}(\bb{z} | \bb{c})$ is then rendered from a desired camera pose $\bb{c}$ by sampling the triplane features and integrating MLP-decoded neural radiance $(\sigma, \bb{f})$ along camera rays. A super-resolution module is followed to modulate the feature map and synthesize the final RGB images at high resolution. We train $G$ and a dual discriminator $D$ with adversarial training.


\begin{figure}[t]
    \centering
    \includegraphics[width=0.5\textwidth]{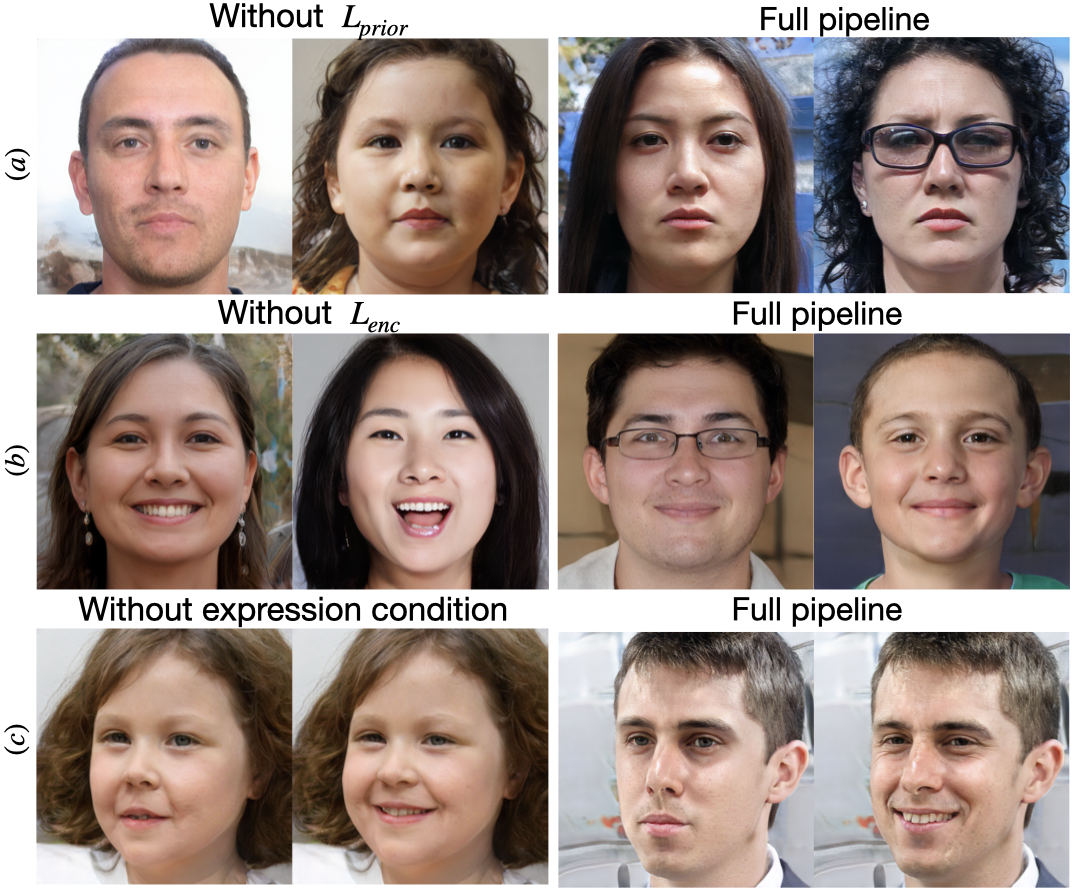}
    \caption{We synthesize different identities with the same shape and expression code respectively in (a) and (b). We observe less shape variation with our geometric prior loss (a) and more consistent expression with our control loss (b). In (c), we show our expression-conditioned NeRF modeling builds rich dynamic details as the subject varies expressions.}
    \label{fig:ablation}
    \vspace{-0.18in}
\end{figure}

\subsection {Controllable 3D-Aware Image Synthesis}
To synthesize an image $\bb{I}(\bb{z} | \bb{c}, \bb{p})$ with desired FLAME control $\bb{p}$, we leverage the EG3D framework to generate a triplane-based 3D volumetric feature space $\cal{M}(\bb{z})$ of a synthesized identity with canonical shape and expression. Guided by our pretrained volumetric correspondence map,
we deform the synthesized feature volume into our target observation space, which is further decoded and volume rendered into high-fidelity head appearance and geometry with the target shape and expression. Our design explicitly disentangles the underlying geometric variations of changing shape and expression from canonical geometry and appearance synthesis. Following EG3D~\cite{chan2021efficient}, we associate each training image with a set of camera parameters $\bb{c}$ and control parameters $\bb{p}$, which are obtained from a nonlinear 2D landmarks-based optimization.

\subsubsection{Semantic Signed Distance Function}
\label{sec:imflame}
\begin{figure}[t]
    \centering
    \includegraphics[width=0.35\textwidth]{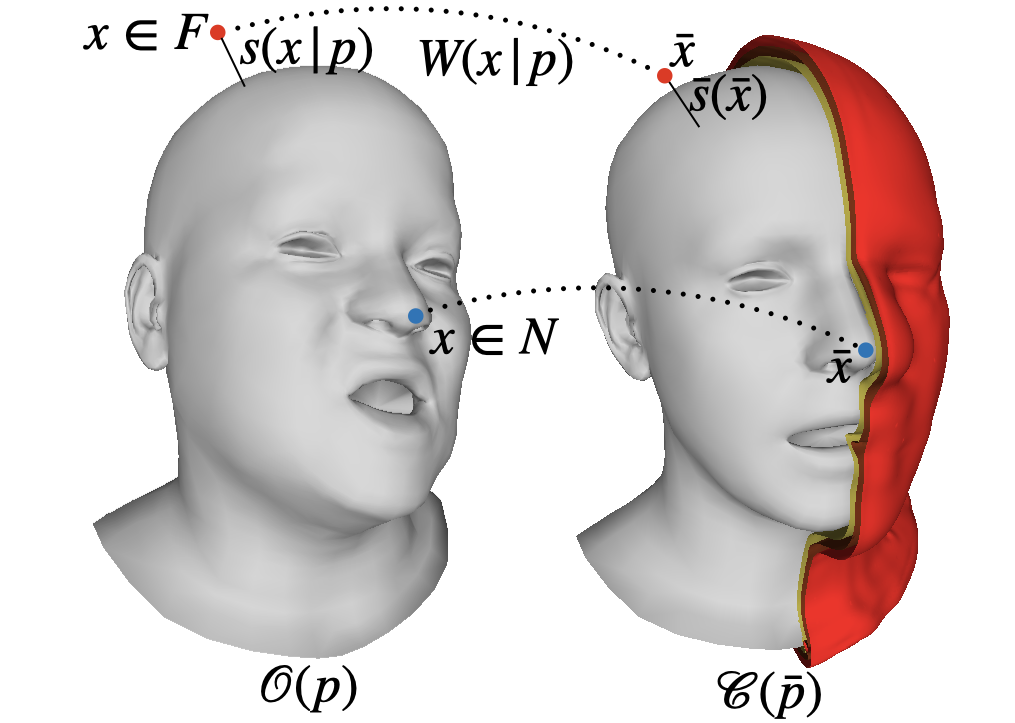}
    \caption{Illustration to semantic SDF learning.}
    \label{fig:illustration}
    \vspace{-0.18in}
\end{figure}

For disentangled geometric modeling, we formulate an implicit semantic SDF representation $W(\bb{x} | \bb{p} = (\bb{\alpha}, \bb{\beta},\bb{\theta}) ) = (s, \bar{\bb{x}})$, where  $\bb{\alpha}, \bb{\beta}$ are the linear shape and expression blendshape coefficients, and $\bb{\theta}$ controls the rotation of a 3-DoF jaw and neck joint. Specifically, given a spatial point $\bb{x}$ in observation space $\cal{O(\bb{p})},$ $W$ returns its 3D correspondence point $\bar{\bb{x}}$ (i.e., semantics) in canonical space $\cal{C(\bb{\bar{p}})}$, with which we project and query the triplane features $\cal{M}(\bb{z}).$ Additionally it also computes the closest signed distance  $s(\bb{x} | \bb{p})$ to the  FLAME mesh surface $\bb{S}(\bb{p}).$ We illustrate the function in Figure~\ref{fig:illustration}. We co-learn the highly-correlated volumetric correspondence and SDF with the property that the signed distance is preserved between canonical and observation correspondence points as $s(\bar{\bb{x}} | \bar{\bb{p}}) = s(\bb{x} | \bb{p})$.


We learn $W(\bb{x} | \bb{p})$ with a large corpus of 3D FLAME meshes $\bb{S}(\bb{p})$ sampled from its parametric control space. Similar to IGR~\cite{icml2020_2086} and imGHUM~\cite{alldieck2021imghum}, we model our implicit field as an MLP and optimize $W(\bb{x} | \bb{p})$ with losses, 
\begin{gather}
L_{iso} = \frac{1}{|N|}\sum_{\bb{x} \in N}(|s(\bb{x} | \bb{p})| + |\nabla s_{\bb{x}}(\bb{x} | \bb{p}) - \bb{n}(\bb{x} | \bb{p})|), \\
L_{eik} = \frac{1}{|F|}\sum_{\bb{x}\in F}\|\nabla s_{\bb{x}}(\bb{x} | \bb{p}) - 1 \|_2, \\
L_{sem} = \frac{1}{|N|}\sum_{\bb{x} \in N}(|\bar{\bb{x}}(\bb{x} | \bb{p}) - \bar{\bb{x}}^*(\bb{x} | \bar{\bb{p}})| )
\end{gather}
where $N, F$ are a batch of on and off surface samples. For the surface samples, the $L_{iso}$ encourages the signed distance values to be on the zero-level-set and the SDF gradient to be equal to the given surface normals $\bb{n}$. The Eikonal loss $L_{eik}$  is derived from~\cite{icml2020_2086} where the SDF is differentiable everywhere with gradient norm 1. The semantic loss $L_{sem}$ supervises the mapping of surface samples $\bb{x} \in F$ to the ground-truth correspondence points $\bar{\bb{x}}^*$ on the canonical
FLAME surface, where $\bar{\bb{x}}^*$ and $\bb{x}$ share the same barycentric coordinates. 
The $L_{eik}$ provides a geometric regularization over the volumetric SDF, whereas the $L_{iso}$ and $L_{sem}$ act as the boundary condition to the SDF and volumetric correspondence field respectively. 

To guide the canonical correspondence learning for off-the-surface points, one could try to minimize the signed distance difference between $s(\bar{\bb{x}} | \bar{\bb{p}})$ and $s(\bb{x} | \bb{p}).$ However, we note that the volume correspondence between the observation and canonical space is still ill-regularized, considering infinite number of points exist with the same signed distance value. We therefore reformulate volumetric correspondence field as $W(\bb{x} | \bb{p}) = (\bar{s}(\bar{\bb{x}}), \bar{\bb{x}}),$ where 
the signed distance of an observation-space point is obtained by mapping it into its canonical correspondence $\bar{\bb{x}}$ and querying the pre-computed canonical SDF $\bar{s}$. Thus we only learn a correspondence field with which we can deform the canonical SDF to different FLAME configurations, and then supervise the deformed SDFs with the FLAME surface boundary, normals ($L_{iso}$) and Eikonal regularization ($L_{eik}$).
As such, even for off-the-surface samples, their canonical correspondences are well regularized in space with the geometric properties of signed distance functions via $L_{iso}$ and $L_{eik}$. In contrast to explicit mesh deformation~\cite{bergman2022generative}, our implicit volumetric correspondence is more accurate, differentiable, smooth everywhere and semantically consistent with the properties of SDF.
 
\begin{figure*}[t]
    \centering
    \includegraphics[width=1.0\textwidth]{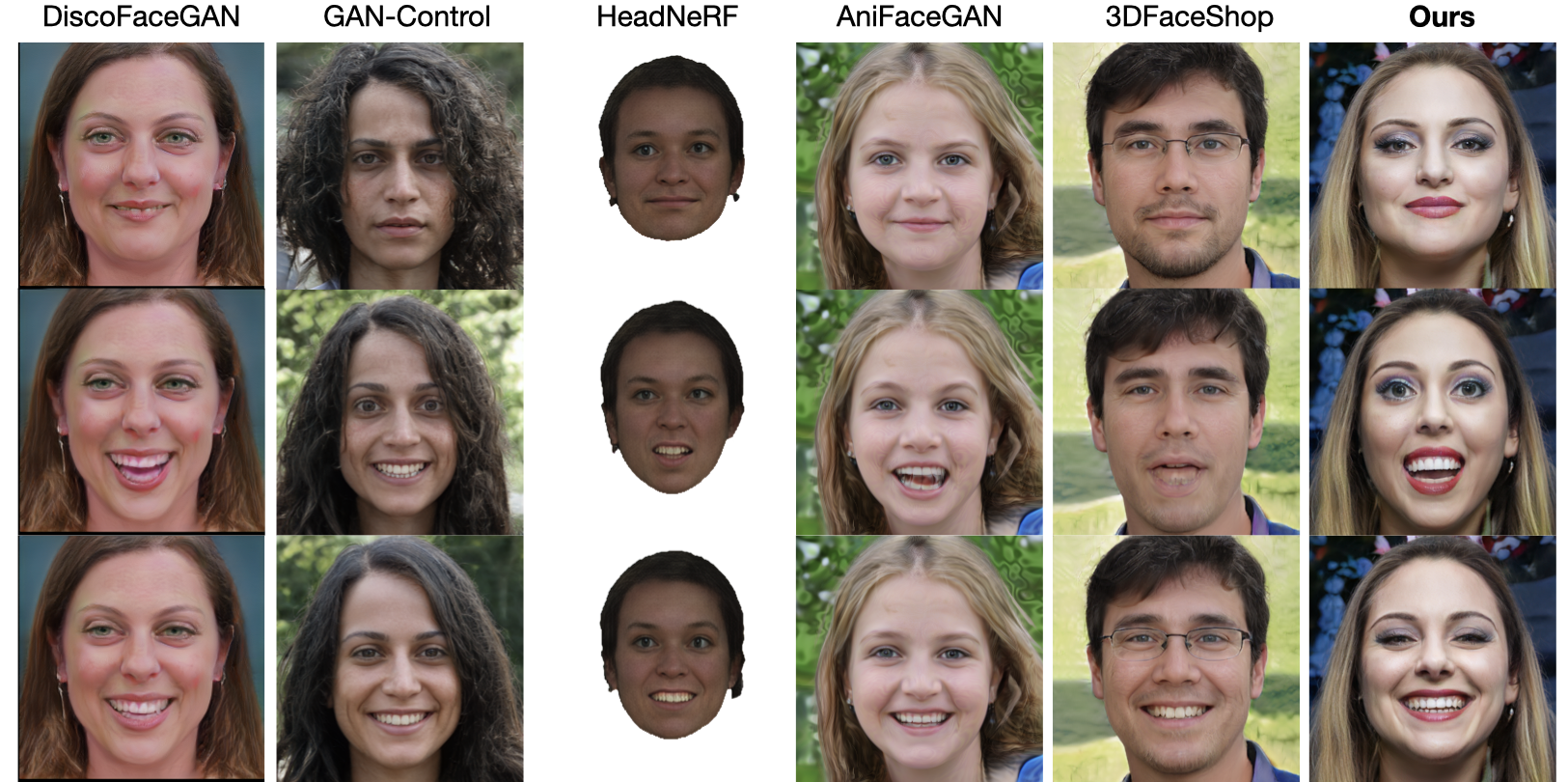}
    \caption{\textbf{Qualitative comparisons in expression control}. Under a similar expression control, our method achieves the best perceptual quality with better identity-preservation (compared to GAN-Control~\cite{shoshan2021gan} and 3DFaceShop~\cite{tang2022explicitly}), more realistic dynamic details (compared to DiscoFaceGAN~\cite{deng2020disentangled}) and higher expression control accuracy (compared to AniFaceGAN~\cite{wu2022anifacegan}, 3DFaceShop~\cite{tang2022explicitly}, HeadNeRF~\cite{hong2022headnerf}).}
    \label{fig:shape_exp_var}
\end{figure*}
\vspace{-0.05in}
\subsubsection{Canonical Generation with Geometric Prior}
\label{sec:canonical}
With our pretrained semantic SDF $W(\bb{x} | \bb{p})$ modeling the shape and expression variation, we leverage the triplane for 3d-aware generation of human heads with canonical shape and expression. In particular, to generate a neural radiance feature $(\sigma, \bb{f})$ for a point $\bb{x}$ in observation space $\cal{O}(\bb{p})$, we use our correspondence function $W$ to back warp $\bb{x}$ into $\bar{\bb{x}}$,  with which we project and sample the canonical triplane features followed with a tiny MLP decoding. 

In spite of the control disentanglement, there is no explicit loss that constrains the triplane-generated neural radiance density to conform to the shape and expression as defined in $\bb{p}.$ Therefore, we guide the generation of neural radiance density field by minimizing its difference to the underlying FLAME head geometry represented with SDF, as 
\begin{gather}
    L_{prior} = \frac{1}{|R|} \sum_{\bb{x}\in R} e^{-\gamma |s(\bb{x}| p)|}| \sigma(\bb{x} | \bb{z}, \bb{p}) - \sigma^{*}(\bb{x} | \bb{p}) |, \\
    \sigma^{*}(\bb{x} | \bb{p}) = \frac{1}{\kappa}\cdot \textrm{Sigmoid}(\frac{-s(\bb{x} | \bb{p})}{\kappa})
\end{gather}
where $R$ is the stratified ray samples for volume rendering and $\kappa$ is a learnable scalar controlling the density tightness around the SDF boundary. Following~\cite{yariv2021volume, or2021stylesdf}, we convert SDF value $s(\bb{x} | \bb{p})$ to proxy 3D density $\sigma^{*}(\bb{x} | \bb{p})$ assuming non-hollow surfaces. We decay the weights for our geometric prior loss $L_{prior}$ as the point moving away from the SDF boundary, allowing higher degrees of freedom in generation of residual geometries, such as hair and glasses. The geometric prior loss effectively guides the 3D head geometry learning but should not be overpowered which otherwise might lead to loss of geometric details. 

\subsubsection{Fine-Grained Expression Control}
\label{sec:supervision}
Our geometric prior loss $L_{prior}$ provides local 3D point-wise guidance, and is able to well regularize the shape generation and achieve coarse-level expression control. However, for delicate expressions, such as eye blinks, $L_{prior}$ provides little supervision as the geometric variation is subtle. Moreover, for regions with complex correspondences, such as around the lips, it is challenging to guide the formation of correct expressions globally, just with point-wise geometric losses. To improve the control granularity, we propose an image-level supervision loss that requires a synthesized image $I_{RGB}(\bb{z} | \bb{c}, \bb{p})$ matching the target expression as defined in the input $\bb{p}$. Using our training images with estimated control labels $\bb{p},$ we first pretrain an image encoder $E(I_{RGB})= (\tilde{\bb{\beta}}, \tilde{\bb{\theta}})$ that regresses the expression coefficients $\tilde{\bb{\beta}}$ and joint poses $\tilde{\bb{\theta}}.$ During our 3D GAN training, we then apply our image-level control supervision as 
\begin{gather}
    L_{enc} = |\tilde{\bb{\beta}} - \bb{\beta}| + |\tilde{\bb{\theta}} - \bb{\theta}| + |\bb{S}(\bb{\alpha}, \tilde{\bb{\beta}}, \tilde{\bb{\theta}}) - \bb{S}(\bb{\alpha}, \bb{\beta}, \bb{\theta})| + \notag\\
    + |\bb{J}\bb{S}(\bb{\alpha}, \tilde{\bb{\beta}}, \tilde{\bb{\theta}}) - \bb{J}\bb{S}(\bb{\alpha}, \bb{\beta}, \bb{\theta})|,
\end{gather}
where $\bb{S}$, $\bb{J}$ are the FLAME mesh vertices and 3D landmarks regressor respectively. While being straightforward for the first 2 terms, the last two terms in $L_{enc}$ penalize deviation of 3D vertex coordinates and surface landmarks after mesh decoding. We note that we do not supervise shape $\bb{\alpha}$ in $L_{enc}$ since our geometric prior loss $L_{prior}$ suffices in shape control already and also due to the ambiguity of shape scaling estimated from monocular images. 

\subsubsection{Dynamic Details Modeling}
\label{sec:dynamic}
To this end, we have achieved controllable static image synthesis. However, one should observe the variation of shading and geometric details in a dynamic head motion, such as the appearance of dimples and eye wrinkles in smiling. We consider the appearance of dynamic details is highly correlated with the driving expression $(\bb{\beta}, \bb{\theta})$. To model such temporal effects, one could try to condition the triplane generation on expression as $\cal{M}(\bb{z}, \bb{\beta}, \bb{\theta}).$ However, such design results in entanglement of expression control with the generative latent code $\bb{z},$ inducing identity or appearance changes when varying expressions. It is also hard to synthesize images out of training distribution of expressions $(\bb{\beta}, \bb{\theta})$. We therefore leave our triplane generation disentangled but when decoding neural feature field $(\sigma, \bb{f})$ from sampled triplane features, we additionally condition the MLP-decoder on $\bb{\beta}$ and $\bb{\theta}.$ Specifically, we have 
\begin{gather}
    (\sigma(\bb{x}), \bb{f}(\bb{x}) | \bb{z}, \bb{p}) = \Phi(\cal{M}(\bar{\bb{x}}(\bb{x} | \bb{p}) | \bb{z}), \phi(\bb{\beta}, \bb{\theta})),
\end{gather}
where both $\Phi$ and $\phi$ are tiny MLPs and $\phi$ regresses an expression-dependent feature vector from  $(\bb{\beta}, \bb{\theta})$ after positional encoding. 
For better extrapolation to novel expressions and jaw poses, we add Gaussian noise to the conditioning parameters to prevent MLP overfitting.

\section{Experiments}
\label{sec:experiments}

\begin{table*}[t]
	\renewcommand{\tabcolsep}{2pt}
	\small
	\begin{subtable}[!t]{0.6\textwidth}
		\centering
	\begin{tabular}{l|cc|cccc}
	\hline
 	Model & FID $\downarrow$  & KID $\downarrow$  & DS$_{\bb{\alpha}}$$\uparrow$ & DS$_{\bb{\beta},\bb{\theta}}$$\uparrow$ & DS$_{\bb{c}}$ $\uparrow$ & IS$\uparrow$  \\
	\hline
	DiscoFaceGAN-SR~\cite{deng2020disentangled} & 42.6 & 0.04 &  - & 27.2 & 0.11 & 0.59\\
	GAN-Control~\cite{shoshan2021gan} & 9.8 & 0.005 &  - & 43.7 & 0.28& 0.62\\
	\hline
	HeadNeRF~\cite{hong2022headnerf}  & 163 & 0.168 &  1.33 & 8.9 & 0.25 & 0.678\\
	AniFaceGAN-SR~\cite{wu2022anifacegan} & 30.5 & 0.024 &  2.0 & 11.1 & 0.04 & 0.73\\
	3DFaceShop~\cite{tang2022explicitly} & 24.8 & 0.018 &  0.45 & 64.2 & 0.029 & 0.594 \\
	\hline
	Ours  & \bb{5.7} & \bb{0.0016} & \bb{3.1} & \bb{71.9} & \bb{0.35} & \bb{0.80} \\
	\hline
	\end{tabular}
		\caption{Baseline comparisons.
		}
	\label{tab:baseline_comp}
	\end{subtable}
	\hspace{\fill}
	\begin{subtable}[!t]{0.4\textwidth}
		\centering
		\vspace{-0.1in}
	\begin{tabular}{l|cc|c}
	\hline
	Model & Ours wo $L_{enc}$  & Ours wo $L_{prior}$ & Ours \\
	\hline
	FID $\downarrow$  & 6.0& 6.1 & \bb{5.7}\\
	KID  $\downarrow$ & 0.00172 &0.00174 & \bb{0.0016}\\
	\hline
	ASD (cm)  $\downarrow$ & \bb{0.124}& 0.135 & \bb{0.124}\\
	$var({\bb{\alpha}})\downarrow$  & 0.000525 & 0.000577 & \bb{0.000515}\\
	AED (cm)  $\downarrow$  &0.1233 & 0.0813 & \bb{0.0797} \\
	$var({\bb{\beta}, \bb{\theta}})\downarrow$  & 0.004707& 0.00406 & \bb{0.00404} \\
	\hline
	\end{tabular}
		\caption{Ablation study.
		}
	\label{tab:ablation}
	\end{subtable}
	\hspace{\fill} 
	\vspace{-0.06in}
	\caption{ (a) Our method outperforms the prior 2D and 3D controllable image synthesis methods in both image quality and control disentanglement. (b) Our method demonstrates the best control accuracy over shape and expression with geometric prior $L_{prior}$ and self-supervised reconstruction loss $L_{enc}.$
	}
	\label{ablations}
	\vspace{-0.15in}
\end{table*}

\paragraph {Training and Dataset.} Our training is devised into two stages. In the pretraining stage, we build our semantic SDF $W$ using a four 192-dimensional MLP with a collection of 150K FLAME meshes by Gaussian sampling of the control parameters $\bb{p}.$ For each mesh, we sample 4K surface samples $N$ with surface normals $\bb{n}$ and ground-truth correspondence point $\bar{\bb{x}}^*$, and 4K off surface samples $F$ distributed uniformly inside the box-bounded volume. 
Our canonical mesh has a neutral shape $\bb{\alpha}=0$, expression $\bb{\beta}=0$ and neck pose, but with an opening jaw at 10 degrees. We close the face connectivity between the lips for a watertight geometry and also as a proxy geometry for the mouth cavity modeling. We open the jaw at the canonical space such that the spatial geometry and appearance inside the mouth can be modeled with distinguishable triplane features.   
In the second training stage, we freeze the weights of $W$ for best efficiency and train our model on the FFHQ~\cite{karras2019style}, a human face dataset with $70,000$ in-the-wild images. For each image, we estimate the camera pose $\bb{c}$ and FLAME parameters $\bb{p}$ with a nonlinear optimization in fitting 2D landmarks, assuming zero root and neck rotation and camera located $1.0$m away from the world origin. We refer to the supplementary materials for more details. We rebalance the FFHQ dataset by duplicating images with large head poses and jaw opening expressions. We note that we also use pairs of images and $(\bb{\beta},\bb{\theta})$ for fine-tuning of a light-weight image encoder $E,$ using a ResNet50~\cite{he2016deep} backbone followed with a single-layer MLP. 

\subsection{Qualitative Comparisons}

\paragraph{Controlled Portrait Synthesis.}
Our framework enables high-fidelity head synthesis with disentangled control over camera pose, shape and expression, as shown in Figure~\ref{fig:teaser}. We compare our method with prior controllable portrait synthesis works include DiscoFaceGAN~\cite{deng2020disentangled}, GAN-Control~\cite{shoshan2021gan}, HeadNeRF~\cite{hong2022headnerf}, AniFaceGAN~\cite{wu2022anifacegan} and 3DFaceShop~\cite{tang2022explicitly}. The first two methods integrates 3DMM controls into 2D face synthesis. HeadNeRF~\cite{hong2022headnerf} is a 3D parametric NeRF-based head model built from both indoor multiview image capture datasets and in-the-wild monocular image collections. AniFaceGAN~\cite{wu2022anifacegan} and 3DFaceShop~\cite{tang2022explicitly} are concurrent 3D-aware controllable GANs in face synthesis. 

 We qualitatively compare with prior work in Figure~\ref{fig:shape_exp_var} for expression control.
GAN-Control shows high-quality expression manipulation but with noticeable identity variation with differences in hair, head shape contour and background, whereas DiscoFaceGAN better maintains identity but with lower perceptual quality, and lacks dynamic details as expression varies. The expression control space of 3DFaceShop~\cite{tang2022explicitly} is sensitive, suffering from appearance changes even with minor expression variation. AniFaceGAN demonstrates visually-consistent expression editing but with limited resolution and image quality, e.g., with blurry artifacts in hair and teeth. HeadNeRF ensures decent consistency in image generation by rendering the conditional NeRF but lacks fine details with very limited perceptual quality. In contrast, our approach produces the most compelling images with consistent appearance and dynamic details under expression changes. 
Moreover, controllable neck pose is a key factor towards realistic video avatar in applications like talking head synthesis. As shown in Figure.~\ref{fig:teaser}, our method achieves explicit control of neck and jaw poses which are seldom explored in prior work. 
Please refer to supplementary material for qualitative comparison in view consistency and shape editing.

\vspace{-0.05in}
\paragraph{Dynamic details.}
Our method presents highly-consistent image synthesis in control of shape, expression and camera poses, but also depicts temporal realism in portrait animation, credited to the modeling of expression-dependent dynamic details. As shown in the last row of Figure~\ref{fig:ablation}, the appearance of wrinkles around the mouth and eyes when transiting from a neural expression to smiling largely enhances the animation realism. In comparison, without explicitly modeling the dynamic details, the wrinkles are embedded in the appearance and do not vary with expressions.  
\vspace{-0.15in}
\paragraph{Expressive Head Synthesis with Extrapolated Controls.}
AniFaceGAN~\cite{wu2022anifacegan} and 3DFaceShop~\cite{tang2022explicitly} depict dynamic details as well since their generation of the neural fields is conditioned on the input expression latent code. However, their designs result in shape and expression entanglement with the appearance generation, as reflected in the identity changes as shown in Figure.~\ref{fig:shape_exp_var}. By embedding the controls in a Gaussian latent space with such as Variation AutoEncoder (VAE), their approaches sacrifice expressiveness. The synthesized image quality is also highly correlated to the distribution of training images with target expressions. 

In contrast, our tri-plane generation is explicitly disentangled from shape and expression control. Moreover, the volume deformation is independently learnt from the deformed FLAME
mesh collections which offer abundant 3D geometric knowledge with largely augmented control space. Therefore we are much less dependent on the distribution of the training images and support better extrapolation to unseen novel expressions. In Figure.~\ref{fig:teaser}, we show high-quality synthesized head with extreme jaw and neck articulated movements which do not exist in our training images. Our expression control is also more expressive, supporting subtle expressions like eye blinks (Figure.~\ref{fig:teaser}~\ref{fig:ablation}).

\subsection{Quantitative Comparisons}
\paragraph{Image Quality}
We measure the image quality with Frechet Inception Distance (FID)~\cite{heusel2017gans} and Kernel Inception Distance~\cite{bi2019deep} between 50K  randomly synthesized images and 50K randomly sampled real images at the resolution of $512\times 512$. Since DiscoFaceGAN~\cite{deng2020disentangled} and AniFaceGAN~\cite{wu2022anifacegan} only synthesize images at $256\times256$ resolution, we utilize a state-of-the-art super-resolution model, SwinIR~\cite{liang2021swinir} to upsample into $512\times512$ for a fair comparison. As shown in Table.~\ref{tab:baseline_comp}, our method is superior in both FID and KID than all prior work, demonstrating the most compelling image quality. We note that the original EG3D has a slightly lower FID at $4.8$ which is expected since we introduce controllability with additional loss regularization. 
\vspace{-0.25in}
\paragraph{Disentanglement}
To evaluate the disentangled controllability of our model  over shape, expression and camera pose, we measure the disentanglement score~\cite{deng2020disentangled} of synthesized images as DS$_{\bb{\alpha}}$, DS$_{\bb{\beta},\bb{\theta}}$ and DS$_{\bb{c}}$ respectively. DS measures the stability of other factors when a single attribute is modified in image synthesis. We employ DECA~\cite{DECA:Siggraph2021} for estimation of FLAME parameters from generated images  and calculate the variance of the estimated parameters $(\bb{\alpha}, \{\bb{\beta},\bb{\theta}\}, \bb{c}).$ Specifically the DS$_i$ is calculated as
\begin{equation}
    DS_i = \prod_{j\neq i} \frac{var(i)}{var(j)}, \,\, i, j \in \{\bb{\alpha}, \{\bb{\beta},\bb{\theta}\}, \bb{c}\}.
\end{equation}
Higher value of DS indicates better disentanglement. Additionally, we evaluate the identity similarity between pairs of synthesized images with random camera poses and expressions by calculating the cosine similarity of the face embeddings with a pre-trained face recognition module~\cite{kim2022adaface}. Our approach demonstrates the best disentanglement numerically over all prior work as indicated in Table.~\ref{tab:baseline_comp}.

\subsection{Ablation Studies}
We ablate the efficacy of the individual component by removing it from our full pipeline. As shown in Figure.~\ref{fig:ablation}, we show the loss of control accuracy over shape and expression respectively when removing  the geometric prior loss $L_{prior}$ and control loss $L_{enc}.$ Conditioning the neural radiance field on expression is also critical to the modeling of dynamic details. Numerically we validate the efficacy of $L_{prior}$ and $L_{enc}$ with Average Shape Distance (ASD) and Average Expression Distance (AED). From the prior distribution, we randomly sample $500$ shapes and expressions, with which we synthesize images with $10$ different identities. We then reconstruct FLAME parameters from the synthesized images and compare against the input control. We compute 3D per-vertex $L_1$ distance of FLAME meshes for ASD while we use 3D landmarks $L_1$ distance for AED calculation. Additionally we compute the variance of estimated shapes and expressions within each control group, with lower value indicating more precise control. The efficacy of $L_{prior}$ and $L_{enc}$ is well evidenced in Table.~\ref{tab:ablation}, with even slight image quality improvements. 


\subsection{Applications}
\paragraph{Talking Head Video Generation.} 
In Figure.~\ref{fig:teaser}, we showcase talking head video generation in controllable views driven with animation sequences of FLAME. Thanks to the high control accuracy, our method is able to synthesize various talking head videos with the same head movements and expressions performed by different identities. 
Our method is expressive in depicting both large articulated neck and jaw movements and subtle facial expressions like eye blinks, with rich dynamic details. Shape manipulation is easily achievable as well by modifying the shape parameters. Please refer to our supplementary materials for more results in high resolution. 
\vspace{-0.1in}
\paragraph{Portrait Image Manipulation and Animation.}
As illustrated in Figure.~\ref{fig:teaser}, our model also supports 3D-aware face reenactment of a single-view portrait to a video sequence. To achieve that, we perform an optimization in the latent Z+ space~\cite{karras2019style} to find the corresponding latent embedding, with FLAME parameter and camera pose estimated from the input portrait. With a frozen generator, the optimization is performed  by measuring the similarity between generated image and real image using the $L_2$ loss and LPIPS loss~\cite{zhang2018unreasonable}. For better reconstruction quality, we alter the parameters of the tri-plane synthesis module with a fixed optimized latent code~\cite{roich2022pivotal}. After that one can explicitly manipulate the portrait with a preserved identity and in a different camera pose and expression. With the expression codes $(\bb{\beta},\bb{\theta})$ reconstructed from a video sequence, we are also able to reenact the portrait to the video motion. 
\vspace{-0.1in}
\paragraph{Societal Impact.} Our work focuses on improving the controllability of 3D-aware GANs in technical aspects and is not specifically designed for any malicious uses. This being said, we do see that the method could be potentially extended into controversial applications such as generating fake videos. Therefore, we believe that the synthesized images and videos should present themselves as synthetic.

\section{Conclusion}
\label{sec:conclusion}
In this work, we introduce OmniAvatar, a novel 3D-aware generative model for synthesis of controllable high-fidelity human head. Our model achieves disentangled semantic control by factoring the generative process into 3D-aware canonical head synthesis and implicit volume deformation to target shapes and expressions. By learning a deformable signed distance function with  canonical volumetric correspondence, we instill the geometric prior knowledge of a 3D statistical head model into the generation of neural radiance fields, enabling superior generative capability to novel shapes and expressions. Our approach demonstrates expressive and compelling talking head generation and portrait image animation with fine-grained control accuracy and temporal realism. We believe the proposed method presents an interesting direction for 3D avatar creation and animation, which sheds light on many potential downstream tasks. 

{\small
\bibliographystyle{ieee_fullname}
\bibliography{egbib}

\begin{thebibliography}{10}\itemsep=-1pt

\bibitem{achlioptas2018learning}
Panos Achlioptas, Olga Diamanti, Ioannis Mitliagkas, and Leonidas Guibas.
\newblock Learning representations and generative models for 3d point clouds.
\newblock In {\em {ICML}}, 2018.

\bibitem{alldieck2021imghum}
Thiemo Alldieck, Hongyi Xu, and Cristian Sminchisescu.
\newblock imghum: Implicit generative models of 3d human shape and articulated
  pose.
\newblock In {\em {ICCV}}, pages 5461--5470, 2021.

\bibitem{barron2021mip}
Jonathan~T Barron, Ben Mildenhall, Matthew Tancik, Peter Hedman, Ricardo
  Martin-Brualla, and Pratul~P Srinivasan.
\newblock Mip-nerf: A multiscale representation for anti-aliasing neural
  radiance fields.
\newblock In {\em iccv}, pages 5855--5864, 2021.

\bibitem{bergman2022generative}
Alexander~W Bergman, Petr Kellnhofer, Yifan Wang, Eric~R Chan, David~B Lindell,
  and Gordon Wetzstein.
\newblock Generative neural articulated radiance fields.
\newblock {\em arXiv preprint arXiv:2206.14314}, 2022.

\bibitem{bi2019deep}
Sai Bi, Kalyan Sunkavalli, Federico Perazzi, Eli Shechtman, Vladimir~G Kim, and
  Ravi Ramamoorthi.
\newblock Deep cg2real: Synthetic-to-real translation via image
  disentanglement.
\newblock In {\em iccv}, pages 2730--2739, 2019.

\bibitem{blanz1999morphable}
Volker Blanz and Thomas Vetter.
\newblock A morphable model for the synthesis of 3d faces.
\newblock In {\em Proceedings of the 26th annual conference on Computer
  graphics and interactive techniques}, 1999.

\bibitem{chan2021efficient}
Eric~R Chan, Connor~Z Lin, Matthew~A Chan, Koki Nagano, Boxiao Pan, Shalini
  De~Mello, Orazio Gallo, Leonidas Guibas, Jonathan Tremblay, Sameh Khamis,
  et~al.
\newblock Efficient geometry-aware 3d generative adversarial networks.
\newblock {\em {CVPR}}, 2022.

\bibitem{chan2021pi}
Eric~R Chan, Marco Monteiro, Petr Kellnhofer, Jiajun Wu, and Gordon Wetzstein.
\newblock pi-gan: Periodic implicit generative adversarial networks for
  3d-aware image synthesis.
\newblock In {\em Proceedings of the IEEE/CVF conference on computer vision and
  pattern recognition}, pages 5799--5809, 2021.

\bibitem{chen2022sofgan}
Anpei Chen, Ruiyang Liu, Ling Xie, Zhang Chen, Hao Su, and Jingyi Yu.
\newblock Sofgan: A portrait image generator with dynamic styling.
\newblock {\em sigg}, 41(1):1--26, 2022.

\bibitem{deng2020disentangled}
Yu Deng, Jiaolong Yang, Dong Chen, Fang Wen, and Xin Tong.
\newblock Disentangled and controllable face image generation via 3d
  imitative-contrastive learning.
\newblock In {\em cvpr}, pages 5154--5163, 2020.

\bibitem{deng2021gram}
Yu Deng, Jiaolong Yang, Jianfeng Xiang, and Xin Tong.
\newblock Gram: Generative radiance manifolds for 3d-aware image generation.
\newblock {\em {CVPR}}, 2022.

\bibitem{DECA:Siggraph2021}
Yao Feng, Haiwen Feng, Michael~J. Black, and Timo Bolkart.
\newblock Learning an animatable detailed {3D} face model from in-the-wild
  images.
\newblock In {\em {ACM Transactions on Graphics}}, volume~40, 2021.

\bibitem{gafni2021dynamic}
Guy Gafni, Justus Thies, Michael Zollhofer, and Matthias Nie{\ss}ner.
\newblock Dynamic neural radiance fields for monocular 4d facial avatar
  reconstruction.
\newblock In {\em Proceedings of the IEEE/CVF Conference on Computer Vision and
  Pattern Recognition}, pages 8649--8658, 2021.

\bibitem{ghosh2020gif}
Partha Ghosh, Pravir~Singh Gupta, Roy Uziel, Anurag Ranjan, Michael~J Black,
  and Timo Bolkart.
\newblock Gif: Generative interpretable faces.
\newblock In {\em 3dv}, pages 868--878. IEEE, 2020.

\bibitem{goodfellow2014generative}
Ian Goodfellow, Jean Pouget-Abadie, Mehdi Mirza, Bing Xu, David Warde-Farley,
  Sherjil Ozair, Aaron Courville, and Yoshua Bengio.
\newblock Generative adversarial nets.
\newblock In {\em {NeurIPS}}, 2014.

\bibitem{icml2020_2086}
Amos Gropp, Lior Yariv, Niv Haim, Matan Atzmon, and Yaron Lipman.
\newblock Implicit geometric regularization for learning shapes.
\newblock In {\em icml}, pages 3789--3799, 2020.

\bibitem{gu2021stylenerf}
Jiatao Gu, Lingjie Liu, Peng Wang, and Christian Theobalt.
\newblock Stylenerf: A style-based 3d-aware generator for high-resolution image
  synthesis.
\newblock {\em {CVPR}}, 2022.

\bibitem{guo2021ad}
Yudong Guo, Keyu Chen, Sen Liang, Yong-Jin Liu, Hujun Bao, and Juyong Zhang.
\newblock Ad-nerf: Audio driven neural radiance fields for talking head
  synthesis.
\newblock In {\em iccv}, pages 5784--5794, 2021.

\bibitem{he2016deep}
Kaiming He, Xiangyu Zhang, Shaoqing Ren, and Jian Sun.
\newblock Deep residual learning for image recognition.
\newblock In {\em cvpr}, pages 770--778, 2016.

\bibitem{heusel2017gans}
Martin Heusel, Hubert Ramsauer, Thomas Unterthiner, Bernhard Nessler, and Sepp
  Hochreiter.
\newblock Gans trained by a two time-scale update rule converge to a local nash
  equilibrium.
\newblock {\em {NeurIPS}}, 2017.

\bibitem{hong2022headnerf}
Yang Hong, Bo Peng, Haiyao Xiao, Ligang Liu, and Juyong Zhang.
\newblock Headnerf: A real-time nerf-based parametric head model.
\newblock In {\em cvpr}, pages 20374--20384, 2022.

\bibitem{kania2022conerf}
Kacper Kania, Kwang~Moo Yi, Marek Kowalski, Tomasz Trzci{\'n}ski, and Andrea
  Tagliasacchi.
\newblock Conerf: Controllable neural radiance fields.
\newblock In {\em cvpr}, pages 18623--18632, 2022.

\bibitem{karras2021alias}
Tero Karras, Miika Aittala, Samuli Laine, Erik H{\"a}rk{\"o}nen, Janne
  Hellsten, Jaakko Lehtinen, and Timo Aila.
\newblock Alias-free generative adversarial networks.
\newblock {\em {NeurIPS}}, 2021.

\bibitem{karras2019style}
Tero Karras, Samuli Laine, and Timo Aila.
\newblock A style-based generator architecture for generative adversarial
  networks.
\newblock In {\em {CVPR}}, 2019.

\bibitem{Karras2020stylegan2}
Tero Karras, Samuli Laine, Miika Aittala, Janne Hellsten, Jaakko Lehtinen, and
  Timo Aila.
\newblock Analyzing and improving the image quality of {StyleGAN}.
\newblock In {\em {CVPR}}, 2020.

\bibitem{kim2022adaface}
Minchul Kim, Anil~K Jain, and Xiaoming Liu.
\newblock Adaface: Quality adaptive margin for face recognition.
\newblock In {\em cvpr}, pages 18750--18759, 2022.

\bibitem{kowalski2020config}
Marek Kowalski, Stephan~J Garbin, Virginia Estellers, Tadas Baltru{\v{s}}aitis,
  Matthew Johnson, and Jamie Shotton.
\newblock Config: Controllable neural face image generation.
\newblock In {\em eccv}, pages 299--315. Springer, 2020.

\bibitem{li2019pu}
Ruihui Li, Xianzhi Li, Chi-Wing Fu, Daniel Cohen-Or, and Pheng-Ann Heng.
\newblock Pu-gan: a point cloud upsampling adversarial network.
\newblock In {\em {ICCV}}, 2019.

\bibitem{FLAME:SiggraphAsia2017}
Tianye Li, Timo Bolkart, Michael.~J. Black, Hao Li, and Javier Romero.
\newblock Learning a model of facial shape and expression from {4D} scans.
\newblock {\em {ACM Transactions on Graphics}}, 36(6):194:1--194:17, 2017.

\bibitem{liang2021swinir}
Jingyun Liang, Jiezhang Cao, Guolei Sun, Kai Zhang, Luc Van~Gool, and Radu
  Timofte.
\newblock Swinir: Image restoration using swin transformer.
\newblock In {\em iccv}, pages 1833--1844, 2021.

\bibitem{Liao2020CVPR}
Yiyi Liao, Katja Schwarz, Lars Mescheder, and Andreas Geiger.
\newblock Towards unsupervised learning of generative models for {3D}
  controllable image synthesis.
\newblock In {\em {CVPR}}, 2020.

\bibitem{liu2021neural}
Lingjie Liu, Marc Habermann, Viktor Rudnev, Kripasindhu Sarkar, Jiatao Gu, and
  Christian Theobalt.
\newblock Neural actor: Neural free-view synthesis of human actors with pose
  control.
\newblock {\em {ACM Trans. on Graphics}}, 2021.

\bibitem{mildenhall2020nerf}
Ben Mildenhall, Pratul~P Srinivasan, Matthew Tancik, Jonathan~T Barron, Ravi
  Ramamoorthi, and Ren Ng.
\newblock Nerf: Representing scenes as neural radiance fields for view
  synthesis.
\newblock In {\em {ECCV}}, 2020.

\bibitem{hologan}
Thu Nguyen-Phuoc, Chuan Li, Lucas Theis, Christian Richardt, and Yong-Liang
  Yang.
\newblock {HoloGAN}: {U}nsupervised learning of {3D} representations from
  natural images.
\newblock In {\em {ICCV}}, 2019.

\bibitem{nguyen2020blockgan}
Thu Nguyen-Phuoc, Christian Richardt, Long Mai, Yong-Liang Yang, and Niloy
  Mitra.
\newblock {BlockGAN}: Learning {3D} object-aware scene representations from
  unlabelled images.
\newblock In {\em {NeurIPS}}, 2020.

\bibitem{niemeyer2021giraffe}
Michael Niemeyer and Andreas Geiger.
\newblock Giraffe: Representing scenes as compositional generative neural
  feature fields.
\newblock In {\em {CVPR}}, 2021.

\bibitem{or2021stylesdf}
Roy Or-El, Xuan Luo, Mengyi Shan, Eli Shechtman, Jeong~Joon Park, and Ira
  Kemelmacher-Shlizerman.
\newblock Stylesdf: High-resolution 3d-consistent image and geometry
  generation.
\newblock {\em {CVPR}}, 2022.

\bibitem{park2021nerfies}
Keunhong Park, Utkarsh Sinha, Jonathan~T Barron, Sofien Bouaziz, Dan~B Goldman,
  Steven~M Seitz, and Ricardo Martin-Brualla.
\newblock Nerfies: Deformable neural radiance fields.
\newblock In {\em {ICCV}}, 2021.

\bibitem{park2021hypernerf}
Keunhong Park, Utkarsh Sinha, Peter Hedman, Jonathan~T Barron, Sofien Bouaziz,
  Dan~B Goldman, Ricardo Martin-Brualla, and Steven~M Seitz.
\newblock Hypernerf: A higher-dimensional representation for topologically
  varying neural radiance fields.
\newblock {\em {ACM Transactions on Graphics}}, 2022.

\bibitem{paysan20093d}
Pascal Paysan, Reinhard Knothe, Brian Amberg, Sami Romdhani, and Thomas Vetter.
\newblock A 3d face model for pose and illumination invariant face recognition.
\newblock In {\em 2009 sixth IEEE international conference on advanced video
  and signal based surveillance}, pages 296--301. Ieee, 2009.

\bibitem{piao2021inverting}
Jingtan Piao, Keqiang Sun, Quan Wang, Kwan-Yee Lin, and Hongsheng Li.
\newblock Inverting generative adversarial renderer for face reconstruction.
\newblock In {\em cvpr}, pages 15619--15628, 2021.

\bibitem{pumarola2021d}
Albert Pumarola, Enric Corona, Gerard Pons-Moll, and Francesc Moreno-Noguer.
\newblock D-nerf: Neural radiance fields for dynamic scenes.
\newblock In {\em {CVPR}}, 2021.

\bibitem{roich2022pivotal}
Daniel Roich, Ron Mokady, Amit~H Bermano, and Daniel Cohen-Or.
\newblock Pivotal tuning for latent-based editing of real images.
\newblock {\em ACM Transactions on Graphics (TOG)}, 42(1):1--13, 2022.

\bibitem{schwarz2020graf}
Katja Schwarz, Yiyi Liao, Michael Niemeyer, and Andreas Geiger.
\newblock Graf: Generative radiance fields for 3d-aware image synthesis.
\newblock {\em {NeurIPS}}, 2020.

\bibitem{shi2021lifting}
Yichun Shi, Divyansh Aggarwal, and Anil~K Jain.
\newblock Lifting 2d stylegan for 3d-aware face generation.
\newblock In {\em cvpr}, pages 6258--6266, 2021.

\bibitem{shi2021SemanticStyleGAN}
Yichun Shi, Xiao Yang, Yangyue Wan, and Xiaohui Shen.
\newblock Semanticstylegan: Learning compositional generative priors for
  controllable image synthesis and editing.
\newblock In {\em {CVPR}}, 2022.

\bibitem{shoshan2021gan}
Alon Shoshan, Nadav Bhonker, Igor Kviatkovsky, and Gerard Medioni.
\newblock Gan-control: Explicitly controllable gans.
\newblock In {\em Proceedings of the IEEE/CVF International Conference on
  Computer Vision}, pages 14083--14093, 2021.

\bibitem{epigraf}
Ivan Skorokhodov, Sergey Tulyakov, Yiqun Wang, and Peter Wonka.
\newblock Epigraf: Rethinking training of 3d gans.
\newblock {\em arXiv preprint arXiv:2206.10535}, 2022.

\bibitem{sun2022controllable}
Keqiang Sun, Shangzhe Wu, Zhaoyang Huang, Ning Zhang, Quan Wang, and HongSheng
  Li.
\newblock Controllable 3d face synthesis with conditional generative occupancy
  fields.
\newblock {\em arXiv preprint arXiv:2206.08361}, 2022.

\bibitem{Szabo:2019}
Attila Szab\'o, Givi Meishvili, and Paolo Favaro.
\newblock Unsupervised generative {3D} shape learning from natural images.
\newblock {\em arXiv}, 2019.

\bibitem{tang2022explicitly}
Junshu Tang, Bo Zhang, Binxin Yang, Ting Zhang, Dong Chen, Lizhuang Ma, and
  Fang Wen.
\newblock Explicitly controllable 3d-aware portrait generation.
\newblock {\em arXiv preprint arXiv:2209.05434}, 2022.

\bibitem{tewari2020pie}
Ayush Tewari, Mohamed Elgharib, Florian Bernard, Hans-Peter Seidel, Patrick
  P{\'e}rez, Michael Zollh{\"o}fer, and Christian Theobalt.
\newblock Pie: Portrait image embedding for semantic control.
\newblock {\em sigg}, 39(6):1--14, 2020.

\bibitem{tewari2020stylerig}
Ayush Tewari, Mohamed Elgharib, Gaurav Bharaj, Florian Bernard, Hans-Peter
  Seidel, Patrick P{\'e}rez, Michael Zollhofer, and Christian Theobalt.
\newblock Stylerig: Rigging stylegan for 3d control over portrait images.
\newblock In {\em cvpr}, pages 6142--6151, 2020.

\bibitem{tov2021designing}
Omer Tov, Yuval Alaluf, Yotam Nitzan, Or Patashnik, and Daniel Cohen-Or.
\newblock Designing an encoder for stylegan image manipulation.
\newblock {\em sigg}, 40(4):1--14, 2021.

\bibitem{tretschk2021non}
Edgar Tretschk, Ayush Tewari, Vladislav Golyanik, Michael Zollh{\"o}fer,
  Christoph Lassner, and Christian Theobalt.
\newblock Non-rigid neural radiance fields: Reconstruction and novel view
  synthesis of a dynamic scene from monocular video.
\newblock In {\em {ICCV}}, 2021.

\bibitem{wang2021learning}
Ziyan Wang, Timur Bagautdinov, Stephen Lombardi, Tomas Simon, Jason Saragih,
  Jessica Hodgins, and Michael Zollhofer.
\newblock Learning compositional radiance fields of dynamic human heads.
\newblock In {\em cvpr}, pages 5704--5713, 2021.

\bibitem{wu2016learning}
Jiajun Wu, Chengkai Zhang, Tianfan Xue, William~T. Freeman, and Joshua~B.
  Tenenbaum.
\newblock Learning a probabilistic latent space of object shapes via {3D}
  generative-adversarial modeling.
\newblock In {\em {NeurIPS}}, 2016.

\bibitem{wu2022anifacegan}
Yue Wu, Yu Deng, Jiaolong Yang, Fangyun Wei, Qifeng Chen, and Xin Tong.
\newblock Anifacegan: Animatable 3d-aware face image generation for video
  avatars.
\newblock {\em arXiv preprint arXiv:2210.06465}, 2022.

\bibitem{xie2022neural}
Yiheng Xie, Towaki Takikawa, Shunsuke Saito, Or Litany, Shiqin Yan, Numair
  Khan, Federico Tombari, James Tompkin, Vincent Sitzmann, and Srinath Sridhar.
\newblock Neural fields in visual computing and beyond.
\newblock In {\em Computer Graphics Forum}, volume~41, pages 641--676. Wiley
  Online Library, 2022.

\bibitem{xu2021h}
Hongyi Xu, Thiemo Alldieck, and Cristian Sminchisescu.
\newblock H-nerf: Neural radiance fields for rendering and temporal
  reconstruction of humans in motion.
\newblock {\em {NeurIPS}}, 2021.

\bibitem{xu2021generative}
Xudong Xu, Xingang Pan, Dahua Lin, and Bo Dai.
\newblock Generative occupancy fields for 3d surface-aware image synthesis.
\newblock {\em {NeurIPS}}, 34:20683--20695, 2021.

\bibitem{xue2022giraffe}
Yang Xue, Yuheng Li, Krishna~Kumar Singh, and Yong~Jae Lee.
\newblock Giraffe hd: A high-resolution 3d-aware generative model.
\newblock In {\em {CVPR}}, 2022.

\bibitem{yariv2021volume}
Lior Yariv, Jiatao Gu, Yoni Kasten, and Yaron Lipman.
\newblock Volume rendering of neural implicit surfaces.
\newblock In {\em {NeurIPS}}, 2021.

\bibitem{zhang2022avatargen}
Jianfeng Zhang, Zihang Jiang, Dingdong Yang, Hongyi Xu, Yichun Shi, Guoxian
  Song, Zhongcong Xu, Xinchao Wang, and Jiashi Feng.
\newblock Avatargen: a 3d generative model for animatable human avatars.
\newblock {\em arXiv preprint arXiv:2208.00561}, 2022.

\bibitem{zhang2018unreasonable}
Richard Zhang, Phillip Isola, Alexei~A Efros, Eli Shechtman, and Oliver Wang.
\newblock The unreasonable effectiveness of deep features as a perceptual
  metric.
\newblock In {\em cvpr}, pages 586--595, 2018.

\bibitem{zheng2022avatar}
Yufeng Zheng, Victoria~Fern{\'a}ndez Abrevaya, Marcel~C B{\"u}hler, Xu Chen,
  Michael~J Black, and Otmar Hilliges.
\newblock Im avatar: Implicit morphable head avatars from videos.
\newblock In {\em cvpr}, pages 13545--13555, 2022.

\bibitem{zhou2021CIPS3D}
Peng Zhou, Lingxi Xie, Bingbing Ni, and Qi Tian.
\newblock {{CIPS}}-{{3D}}: A {{3D}}-{{Aware Generator}} of {{GANs Based}} on
  {{Conditionally}}-{{Independent Pixel Synthesis}}.
\newblock {\em arXiv}, 2021.

\bibitem{zhuang2021mofanerf}
Yiyu Zhuang, Hao Zhu, Xusen Sun, and Xun Cao.
\newblock Mofanerf: Morphable facial neural radiance field.
\newblock {\em arXiv preprint arXiv:2112.02308}, 2021.

\end{thebibliography}
}

\end{document}